# Probabilistic State-Dependent Grammars for Plan Recognition

**David V. Pynadath**
Information Sciences Institute
University of Southern California
4676 Admiralty Way, Marina del Rey, CA 90292
pynadath@isi.edu

**Michael P. Wellman**
Artificial Intelligence Laboratory
University of Michigan
1101 Beal Ave, Ann Arbor, MI 48109
wellman@umich.edu

## Abstract

Techniques for plan recognition under uncertainty require a stochastic model of the plan-generation process. We introduce *probabilistic state-dependent grammars* (PSDGs) to represent an agent's plan-generation process. The PSDG language model extends probabilistic context-free grammars (PCFGs) by allowing production probabilities to depend on an explicit model of the planning agent's internal and external state. Given a PSDG description of the plan-generation process, we can then use inference algorithms that exploit the particular independence properties of the PSDG language to efficiently answer plan-recognition queries. The combination of the PSDG language model and inference algorithms extends the range of plan-recognition domains for which practical probabilistic inference is possible, as illustrated by applications in traffic monitoring and air combat.

## 1  INTRODUCTION

The problem of plan recognition is to determine the plan of action underlying an agent's behavior, based on partial observation of its behavior up to the current time. We assume that this behavior is a product of executing some plan, constructed to serve the agent's objectives based on its beliefs.[1] The examples in this paper draw from a scenario in traffic monitoring, where the observed agent is driving a car along a three-lane, one-way highway.

The agent begins in an initial context, consisting of its position along the highway, presence of other cars, etc. Its mental state is comprised by its preferences (e.g., utility function over speed), beliefs (e.g., speedometer reading), and capabilities (e.g., driving ability). We assume the planning process to be some rational procedure based on such a mental state. The generated plan then determines (perhaps with some uncertainty) the actions taken by the agent in the world. In the traffic example, the observed driver may plan high-level maneuvers (e.g., change of lane, pass of another car) that it ultimately executes through low-level driving actions (e.g., turning the steering wheel).

The recognizer uses its observations, in whatever form, to generate hypotheses about which top-level plan or intermediate subplans the agent has selected, or which low-level actions it will perform in the future. The resulting candidates, as well as possible evaluations of their plausibilities, form the basis for decisions on potential interactions with the observed agent. In the traffic example, a recognizing driver could observe another car maneuvering nearby and compute a probability distribution over possible plan interpretations and future actions, all as part of its own (possibly decision-theoretic) maneuver-selection process.

### 1.1  BAYESIAN NETWORKS FOR PLAN RECOGNITION

Modeling the uncertainty inherent in planning domains provides one of the most difficult challenges in plan recognition. Approaches based on first-order logic typically appeal to heuristic criteria to distinguish among possible explanations of observed phenomena (Kautz & Allen, 1986; Lin & Goebel, 1991; Tambe & Rosenbloom, 1996). However, to support general decision making based on such observations, we require an account of the *relative likelihood* of these explanations.

The most comprehensive probabilistic approach to plan recognition constructs *plan recognition Bayesian networks* representing the relationships among events and uses standard network inference algorithms to compute posterior probability distributions over possible interpretations (Charniak & Goldman, 1993). These plan recognition Bayesian networks represent a probability distribution over a particular set of observed events, appropriate for the in-tended domain of story understanding. However, in many real-world agent domains, the complete set of observations is enormous. For instance, in the traffic domain, we observe cars' positions along the highway repeatedly over the course of many minutes, possibly even hours. Eventually, we would be unable to represent the accumulating set of observations within a Bayesian network that would be tractable for inference.

---
[1] We discuss our overall plan-recognition framework elsewhere (Pynadath & Wellman, 1995; Pynadath, 1999).

*Dynamic* Bayesian networks (DBNs) (Kjærulff, 1992) represent only a restricted window of the random variables by using a compact belief state to summarize the past observations. The belief state is sufficiently expressive to support exact inference over variables within the window. However, the generality of the DBN representation still leads to intractable inference in many plan-recognition domains. Methods for approximate inference can answer queries with sufficient precision and efficiency for some domains (Lesh & Allen, 1999), but still take several minutes for inference. We would instead like a more restricted language that supports online inference in answering plan-recognition queries in a matter of seconds, as required in the air combat and traffic domains.

### 1.2 PROBABILISTIC GRAMMARS

Pattern-recognition research provides a possible source for such representations, since plans are descriptions of action *patterns*. Grammatical representations are generative and modular, providing an appealing class of languages for specifying pattern-generation processes. If we can model a given plan-generation process within a particular grammatical formalism, then we can use that formalism's inference techniques to answer plan-recognition queries.

One candidate approach would use a probabilistic context-free grammar (PCFG) (Charniak, 1993; Gonzalez & Thomason, 1978) to represent a distribution over possible action sequences. Existing PCFG parsing algorithms would support a restricted set of plan-recognition queries. Other grammatical models (Black et al., 1992; Schabes & Waters, 1993; Carroll & Weir, 1997; Magerman, 1995) make fewer independence assumptions than do PCFGs (thus supporting a wider class of problem domains), while still supporting efficient parsing algorithms. The typical parsing algorithm produces the conditional probability of a particular symbol (subplan) or parse tree (plan instantiation), given a *complete* terminal sequence (sequence of observations). However, these parsing algorithms are unsuitable for most plan-recognition queries, which occur *during* execution, before the entire sequence is available. In addition, the entire sequence may *never* become available if there are missing observations.

In previous work, we have shown how to generate a Bayesian network to answer these more general queries for a given PCFG (Pynadath & Wellman, 1998). These Bayesian networks suffer the same drawbacks as those in existing plan-recognition research, since they, too, must represent the entire set of observations. However, by borrowing the notion of a compact belief state from DBN inference and by exploiting the specific independence assumptions of the underlying grammatical model, we may be able to identify a belief state compact enough for practical inference, while still supporting exact inference.

## 2 PROBABILISTIC STATE-DEPENDENT GRAMMARS

This section defines the probabilistic *state-dependent* grammar (PSDG), which supports such belief-state inference. The PSDG model adds an explicit representation of state to an underlying PCFG model of plan selection. The state model captures the dependence of plan selection on the planning context, including the agent's beliefs about the environment and its preferences over outcomes. The state model also represents the effects of the agent's planning choices on future states (and, consequently, on future planning choices). This section defines the language model and demonstrates its ability to represent plan generation in certain domains. Section 3 describes a practical inference algorithm that can answer plan-recognition queries based on a PSDG representation of an agent's planning process.

### 2.1 PSDG DEFINITION

A *probabilistic state-dependent grammar* (PSDG) is a tuple $\langle \Sigma, N, S, Q, P, \pi_0, \pi_1 \rangle$, where the disjoint sets $\Sigma$ and $N$ specify the terminal and nonterminal symbols, respectively, with $S \in N$ being the start symbol (as in a PCFG). $P$ is the set of productions, taking the form $X \rightarrow \xi$ $(p)$, with $X \in N$, $\xi \in (\Sigma \cup N)^+$ and $p$ the probability that $X$ is expanded into the string $\xi$. $Q^t$ is a time-indexed random variable representing a state space (beyond the grammatical symbols) with domain $Q$.

The PSDG production probability, $p : Q \rightarrow [0, 1]$, is a function of the state. Each production $X \rightarrow \xi$ $(p)$ denotes that the conditional probability of expanding $X$ into the sequence $\xi$, given that the current state $Q^t = q$, is $p(q)$. We specify the time, $t$, of a particular nonterminal symbol as the position of its leftmost descendant terminal symbol within the overall terminal string (where $t = 1$ is the first terminal symbol). We can then define the current state for expanding a symbol at time $t$ as $Q^{t-1}$. For each nonterminal symbol $X \in N$, we require that $\sum_\ell p_\ell(q) = 1$ for all states $q \in Q$, where $p_\ell$ ranges over all the production probability functions for expansions of $X$.

The PSDG function $\pi_0$ specifies the distribution over the initial values of the state variable $Q$, i.e., $\Pr(Q^0 = q) = \pi_0(q)$. The function $\pi_1(q_{t-1}, x, q_t)$ specifies the probability that $Q^t = q_t$ given that $Q^{t-1} = q_{t-1}$ and the terminal

```
0) Drive  →  Stay Drive     (p_0(q) = ···)
1) Drive  →  Left Drive
```
$$p_1(q) = \begin{cases} 0 & \text{if Lane}(q) = \text{left-lane} \\ \cdots & \end{cases}$$
```
2) Drive  →  Right Drive    (p_2(q))
3) Drive  →  Pass Drive     (p_3(q))
4) Drive  →  Exit           (p_4(q))
5) Pass   →  Left Right     (p_5(q))
6) Pass   →  Right Left     (p_6(q))
```

Figure 1: PSDG representation of simplified traffic domain.

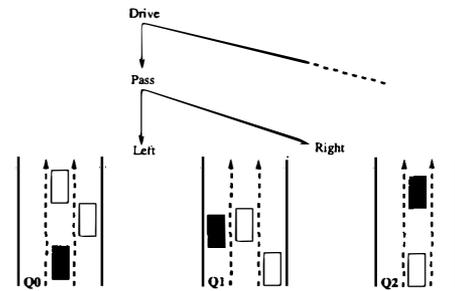

Figure 2: Simple PSDG parse tree from traffic domain.

symbol chosen in the intervening interval is $x$. The value of $Q^t$ is conditionally independent of past values of $Q$ (as well as all symbols in the parse tree selected before $t$) given the value of $Q^{t-1}$ and the terminal symbol chosen in the interval between $t-1$ and $t$.

We can often simplify a PSDG domain by viewing the state as a conjunction of somewhat orthogonal features representing individual aspects of the context. Production probabilities are functions of only those features that influence the choice in expanding a particular symbol. For instance, a driver's decision to accelerate or decelerate may depend on only its preferred traveling speed and the current speed of the car in front, without depending on the location of the intended exit. Likewise, the distribution over a particular feature can depend on other certain features, without having to depend on all (e.g., the driver's position along the highway changes with the current speed, but it does not depend on the current lane). Whereas we often refer to the state as a single variable for intelligibility, Section 3's inference algorithms do exploit factored state representations.

## 2.2 SIMPLE PSDG EXAMPLE

Consider the PSDG of Figure 1, representing a simplified generative model of driving plans. The state includes the observable features of the driver's position and speed, as well as the positions and speeds of other cars on the highway. The state also includes aspects of the driver's mental state, such as the agent's preferences about driving speed, distance from other cars, intended exit, etc. Figure 2 presents one possible instance of the agent's plan generation and execution, as an illustration of PSDG parse tree generation. The picture labeled $Q^0$ in the bottom left corner of the diagram represents the observable portion of the initial state. The solid black rectangle is the driver whose planning process we are trying to recognize. The white rectangles are the other cars on the highway.

To clarify the specification of certain plan events important for describing both generation and recognition, we define random variables, $N_\ell^t$ (nonterminals), $P_\ell^t$ (productions), and $\Sigma^t$ (terminals), to represent an entire path from root node to leaf node (i.e., the stack of active plans). The $t$ index indicates time from left to right in the parse tree.

The $\ell$ index represents the distance of a given symbol from the root node (always the start symbol). The root node has $\ell = 1$, and all other symbol nodes have an $\ell$ index of one more than their parent nodes. In the example, the driver's plan originates with the start symbol, $N_1^1 =$ Drive.

The driver then chooses among the five possible expansions shown in Figure 1. The production variable, $P_1^1$, indicates the production chosen, as well as what symbol on the right-hand side is currently being expanded. In the parse tree of Figure 2, the driver selects the production Drive → Pass Drive with probability $p_3(Q^0)$. Therefore, $P_1^1 = \langle 3, 1 \rangle$, where the first number is the production index, and the second indicates that the currently active child symbol is the first symbol on the right-hand side, Pass.

The production probability, $p_3(Q^0)$, summarizes the plan-selection process conditioned on the context, $Q_0$. We can view the set of probability functions for all expansions of Drive as a decision tree, with the state features as the inputs. For instance, in states where the driver's preferred speed is slower than the current speed of the car in front, passing is very unlikely. Likewise, if the driver's position along the highway is close to the intended exit, then passing is again unlikely, although passing becomes more likely if the driver is of an "aggressive" type.

Given that we have chosen production 3, we continue the plan expansion with the selected child, $N_2^1 =$ Pass. We compute $p_5(Q^0)$ and $p_6(Q^0)$ to determine the probability of passing on the left versus passing on the right. In the example, the driver has chosen to pass on the left ($P_2^1 = \langle 5, 1 \rangle$), so it first executes a Left action.

The random variable $\Sigma^t$ represents the terminal symbol at position $t$ in the final sequence, so in this case, $\Sigma^1 =$ Left. The values of the symbol variables (both terminal and non-terminal) are completely determined given the values of the production variables above them in the hierarchy. In particular, suppose that $P_\ell^t = \langle a, b \rangle$, where production $a$ is $X \to Y_1 Y_2 \cdots Y_m$. In such a case, if $Y_b$ is a terminal symbol, then $\Sigma^t = Y_b$; otherwise, $N_{\ell+1}^t = Y_b$.

Having reached a leaf node for time 1, we can apply the state transition probability, $\pi_1(Q^{t-1}, \Sigma^t, Q^t)$, to compute a distribution over possible values of $Q^1$. This transition probability encodes the world dynamics, including the effects of the observed agent's actions on the state. For instance, the transition probability represents the (possibly uncertain) effect that making a left lane change has on the car's lane position. It also represents the possible changes in the positions of the other cars. The diagram shows one possible value where the driver has moved into the leftmost lane (as a result of selecting the Left action) and moved beyond the other two cars.

The expansion of the top-level plan, Drive, did not complete at time 1, so $N_1^2$ =Drive. The expansion of Pass did not complete either, so $P_1^2 = \langle 3, 1 \rangle$ and $N_2^2$ =Pass. However, at the next level, the terminal symbol Left completed execution in time 1, so we move on to the next symbol to be expanded at that level: $P_2^2 = \langle 5, 2 \rangle$ and $\Sigma^2$ =Right. If any new nonterminal symbols had arisen in this branch at time 2, the state $Q^1$ would form the context when expanding them. We determine the next state value, $Q^2$, by applying the state transition probability, $\pi_1(Q^1, \text{Right}, Q^2)$.

## 2.3 RELATIONSHIP OF PSDGs TO PCFGs

Both the traffic and air-combat PSDGs use finite state spaces. For finite state spaces, we can represent a given PSDG distribution with a corresponding PCFG. This equivalent PCFG symbol space contains tuples $\langle q_i, X, q_f \rangle$, indicating that the PSDG symbol $X$ is expanded starting in initial state $q_i$ and ending in final state $q_f$. Given these new symbol sets, we can construct context-free productions such that the probability of a given PSDG parse tree is identical to the corresponding parse tree from this constructed PCFG. However, the constructed PCFG can be larger than the original PSDG by a factor of $|Q|^{m+1}$, where $m$ is the maximum production length.

In general, if we allow the state space to be infinite, then the PSDG generative model can represent language distributions beyond those allowed by the PCFG model. For instance, the language $\{a^x b^y c^x d^y, y > 0\}$ cannot be represented by a context-free grammar, nor a PCFG. However, if we define the state space $Q = \mathcal{Z}^+ \times \mathcal{Z}^+$ to record the values of $x$ and $y$, then we can specify productions and a *deterministic* state transition function to represent the language. In addition, the inference algorithms of Section 3 support "recognition" queries about this generation mechanism. We omit the PSDG constructions for more general languages here, since, although inference on the constructed PSDGs is possible, it is impractical in general.

## 2.4 IMPLEMENTED PSDGs: TRAFFIC AND AIR COMBAT

Regardless of the potential computational advantage, the separation between the plan and state spaces in the PSDG model can provide a more suitable modeling language, since the dependency structure more closely mirrors that of most planning domains. An examination of the implemented PSDG models of the traffic and air-combat domains illustrates the language's specific strengths and weaknesses. Overall, the complete traffic PSDG has 14 nonterminal symbols (plans), 7 terminal symbols (actions), and 15 state features (with the mean state space size being 431 elements). Three of these state features correspond to aspects of the driver's mental state (preferred speed, intended exit, aggressiveness); the rest of the state features are completely observable. There are a total of 40 productions with a mean length of two symbols. We also implemented a PSDG representation for an air combat domain based on an existing specification (Tambe & Rosenbloom, 1996) using Soar productions (Newell, 1990).

### 2.4.1 State Models in PSDGs

For modeling the planning agent's environment and mental state, the PSDG state variable specification supports arbitrarily complex probabilistic dependency structures. We could capture probabilistic sensor models of the uncertain noise present in the agent's beliefs. However, the agent's beliefs are unobservable, and, as Section 3 discusses, the number and size of unobservable state variables have the biggest impact on the complexity costs of inference. We can often model the agent's noisy beliefs within the productions themselves, thus incurring much less inferential cost. For instance, in the expansions of Drive, there is a nonzero probability for passing even when the driver is at the intended exit. This probability captures the possibility that the driver fails to notice the exit, without requiring an explicit state variable for the driver's belief.

However, we cannot model beliefs and preferences that persist throughout the agent's lifetime through the production probabilities, which are evaluated independently for each episode. PSDG state variables cannot represent distributions over arbitrary utility functions in a manner that supports tractable inference. However, if we can model the planning agent's preferences by a finite set of goals (e.g., intended exit) or finite set of utility function classes (e.g., driver aggressiveness), then we can greatly reduce the complexity of the PSDG state variable representation.

### 2.4.2 Plan Model in PSDGs

The PSDG production format also supports plan generation and execution models much more flexible than that of Figure 1. The PSDG of Figure 1 treats the lane change Left as an atomic action, but the complete PSDG for the traffic domain treats it as only an intended plan with two subcomponents, StartLeft and ExecuteLeft. While in the first subplan, the driver is waiting for conditions to become safe before actually changing lanes. The production probability functions for StartLeft evaluate the highway situation of the current observed state, as well as the unobservable mental state (e.g., the driver's degree of cautiousness or aggressiveness). If conditions are unsafe, the production probability of StartLeft → Stay StartLeft is high, as the driver prefers to stay in the current lane and postpone the lane change. If the conditions become safe, the driver stops waiting with StartLeft → MoveLeft, where MoveLeft is an incremental shift (expected to be 1m) to the left. The production probabilities capture this termination condition through the relative likelihoods of the two StartLeft productions and their dependency on the current state.

Once the expansion of StartLeft terminates, the driver then goes on to expand ExecuteLeft, which produces a series of MoveLeft actions until the car is fully within the new lane. However, the driver also has the option of abandoning the lane change as conditions evolve. If the state is such that the driver no longer desires to move into the left lane (e.g., the car in front moves to a different lane), then the productions StartLeft → Stay and ExecuteLeft → Stay take on the highest production probabilities. Thus, the expansion of Left completes within two cycles, and the driver is free to choose a new maneuver.

Although the conditional production probabilities allow great flexibility, the production structure itself does require a total order over subplans. For example, the original Soar specification of the air-combat domain did not impose an order over subplans. However, the conditions on these particular Soar productions implicitly serialize much of the execution, as the pre-conditions of a particular child are achieved only *after* the execution of its sibling. In general, we would unfortunately have to generate PSDG productions for all of the possible sequences of these children.

The production structure also serializes the execution of plans and actions, precluding the possibility of concurrent actions. The traffic PSDG mimics certain forms of concurrency by using subsequences of symbols. For instance, the simplified grammar of Figure 1 includes action symbols for only lateral movements. The complete PSDG has additional symbols corresponding to acceleration maneuvers as well, with these symbols being interleaved with the lateral movement symbols. However, this mechanism is insufficient for general concurrency, where the plans are not separable and do not have orthogonal effects on the state.

### 2.4.3 World Dynamics in PSDGs

The PSDG state transition probabilities can represent any joint distribution over future world states, conditioned on the past state and the low-level action taken. Most of the relationships expressed by the world dynamics in the traffic example are straightforward. For instance, the value of the lateral position at time $t + 1$ will be to the left of its value at time $t$, given an interposing MoveLeft action. There is uncertainty in the exact change in value, as expressed by the probability distribution in the complete PSDG.

However, the state transition probabilities cannot represent the effects of subplan choices on future states. For instance, we cannot explicitly represent the dependency of the driver's turn indicator on its high-level decisions. We instead introduced terminal symbols representing signaling as an additional concurrent action. The state of the turn indicator is completely determined given the signaling action. In general, we cannot afford to add such indirect plan representations for each such state dependent on a high-level plan. It is important to note that it is the *generative* model that does not capture the dependency of the agent's mental state on plan choices. Once we observe evidence, the inference algorithms of Section 3 *do* capture a conditional dependency in updating the recognizer's beliefs about the agent's mental state.

## 3 PSDG INFERENCE

Although we can perform inference on a given PSDG with a finite state space by generating the corresponding PCFG and using PCFG inference algorithms, the explosion in the size of the symbol space can lead to prohibitive costs. In addition, as described in Section 1.2, existing PCFG algorithms cannot handle most plan-recognition queries.

We can potentially perform inference by generating a DBN representation of a PSDG distribution. The definition of the PSDG language model supports an automatic DBN generation algorithm. The resulting DBN supports queries over the symbol, production, and state random variables. Unfortunately, the complexity of DBN inference is likely to be impractical for most PSDGs, where the belief state must represent the entire joint distribution over all possible combinations of state and parse tree branches. For instance, for the complete PSDG representation of the traffic domain, the DBN belief state would have more than $10^{25}$ entries.

This section presents inference algorithms that exploit the particular structure of the PSDG model to answer a set of queries more restricted than that provided by DBNs. These algorithms use a compact belief state (described in Section 3.1) to answer queries based on observations of the state variables. At time $t$, the recognizer observes some or perhaps all of the features of the state, $Q^t$. We represent this evidence by stating that $Q^t \in R^t$, where $R^t \subseteq Q$ represents the set of possible states consistent with the observations. Based on this evidence, the algorithm presented in Section 3.3 computes posterior probabilities over the individual state elements in $R^t$, as well as posterior probabilities over the possible plans and productions that the agent executed at time $t - 1$. The algorithm presented in Section 3.4 computes the posterior probabilities over the plans and productions that the agent will select at time $t$, as well as updating the recognizer's belief state. A pseudocode description of the algorithm is available in an online appendix[2]. Both the pseudocode and proofs of correctness

---

[2] www.isi.edu/~pynadath/Research/PSDG

are available elsewhere as well (Pynadath, 1999).

## 3.1 COMPACT BELIEF STATE FOR INFERENCE

The high connectivity of the DBN belief state arises from its reliance on strict conditional independence as its structuring property. The DBN representation does not exploit the weaker forms of independence present in the PSDG model. To specify these weaker independence conditions, we first define an expansion, $P_\ell^t = \langle X \to Y_1 Y_2 \cdots Y_m, b\rangle$, as *terminating* at time $t$ if and only if $b = m$ and either $Y_m$ is a terminal symbol or the child expansion, $P_{\ell+1}^t$, terminates at time $t$. If we re-examine the relationship of the production variables $P_\ell^t$ on the previous time slice, we notice that there are two possibilities when the expansion $P_\ell^{t-1}$ has not *terminated*. One possibility is that the expansion of child $N_{\ell+1}^{t-1}$ has not terminated, in which case we continue expanding the child at time $t$ as well, and the value of the parent expansion $P_\ell^{t-1}$ does not change. The other possibility is that the child expansion terminated at time $t - 1$, but there are still more children to expand on the right-hand side of the parent expansion: $P_\ell^{t-1} = \langle X \to Y_1 Y_2 \cdots Y_m, b\rangle$, $b < m$. In this case, we move on to the next child, so $P_\ell^t = \langle X \to Y_1 Y_2 \cdots Y_m, b+1\rangle$. In both cases, the relationship is deterministic. If the parent expansion $P_\ell^{t-1}$ *has* terminated in the previous time slice, then we are choosing a new production based on the new left-hand symbol, $N_\ell^t$, and $Q^{t-1}$, *independent of the symbols and productions of the previous time slice*. We use this independence property and determinism inherent in the PSDG model to treat our beliefs about the plan variables, $N_\ell^t$ and $P_\ell^t$, separately at the different levels, $\ell$, of the hierarchy.

In addition, the DBN representation supports the computation of arbitrary conditional probabilities within the current window of random variables. In most problem domains, we never have direct evidence about the agent's plan choices, but rather only about the current state. For instance, in our traffic example, we can observe the position and speed of cars, but we cannot directly observe aspects of the driver's mental state or its subplan choices (e.g., whether it has chosen a passing maneuver).

To support PSDG inference, DBN inference must maintain a distribution over the joint space of all variables within a given time slice. Our specialized inference algorithms instead maintain a much smaller belief state that summarizes this probability distribution by exploiting the independence properties of the PSDG model and the restricted set of PSDG queries. Table 1 lists the probability tables that form $B^t$, the belief state for time $t$, where $\mathcal{E}^t$ represents all evidence ($Q^t \in R^t$) received through time $t$. The belief component, $B_T^t(\ell, q)$, represents a boolean random variable that is true if and only if the expansion of the symbol at level $\ell$ terminates at time $t$. The overall belief state pro-vides a more compact summarization of observations than a probability distribution over the entire joint space.

| Function | | Definition |
|---|---|---|
| $B_Q^t(q)$ | = | $\Pr(Q^{t-1} = q \mid \mathcal{E}^{t-2})$ |
| $B_N^t(\ell, X, q)$ | = | $\Pr(N_\ell^t = X \mid \mathcal{E}^{t-1}, Q^{t-1} = q)$ |
| $B_P^t(\ell, \langle a, b\rangle, q)$ | = | $\Pr(P_\ell^t = \langle a, b\rangle \mid \mathcal{E}^{t-1}, Q^{t-1} = q)$ |
| $B_\Sigma^t(x, q)$ | = | $\Pr(\Sigma^t = x \mid \mathcal{E}^{t-1}, Q^{t-1} = q)$ |
| $B_T^t(\ell, q)$ | = | $\Pr(T_\ell^t \mid \mathcal{E}^{t-1}, Q^{t-1} = q)$ |
| $B_{T\mid N}^t(\ell, X, q)$ | = | $\Pr(T_\ell^t \mid \mathcal{E}^{t-1}, Q^{t-1} = q, N_\ell^t = X)$ |

Table 1: Belief state structure for PSDG inference.

If the productions introduce possible cycles (as in the PSDG of Figure 1), then there is no bound on the length of parse tree branches, so there is an infinite number of possible hierarchy levels (index $\ell$ in the belief state). However, we can still maintain a finite belief state even if we allow recursive productions of the form $X \to Y_1 Y_2 \cdots Y_{m-1} X$, where the $Y_b \neq X$. The $Y_b$ children have $\ell$ indices as originally specified, but the $X$ on the right-hand side now has the *same* $\ell$ index as the $X$ on the left-hand side. Therefore, we expand the $X$ on the right-hand side from $N_\ell^{t+1}$. We choose a new production at $P_\ell^{t+1}$, so we no longer have any record (in the current branch) of how many levels of nesting have taken place. As long as we have no need of this lost information, we can generate a finite belief-state representation of a PSDG with this limited recursion.

The belief state probabilities are indexed by all of the *individual* states $q \in R^t$ consistent with our observations. The specialized PSDG belief state structure has a space complexity of $O(|R^t| \cdot |P| dm)$, where $d$ is the maximum depth of the hierarchy (the largest $\ell$ value) and $m$ is the maximum production length. The fully connected DBN belief state has a space complexity of $O(|Q| \cdot |P|^d m^d)$.

The compact belief state, $B^t$, no longer explicitly stores the conditional probabilities of production variables given the left-hand symbols, nor those of right-hand symbols given the production variables. We can extract these immediately from the probabilities available in the belief state. For instance, we know that the probability of a production state, $\rho = \langle X \to Y_1 \cdots Y_m, b\rangle$, is zero when the symbol $X$ is not present. Thus, $\Pr(P_\ell^t = \rho \mid N_\ell^t = X, \mathcal{E}^t, Q^t = q) = B_P^t(\ell, \rho, q)/B_N^t(\ell, X, q)$. The conditional probabilities of symbols given parent productions is even simpler, because of their deterministic nature. For instance, for nonterminals $Y_b \in N$, $\Pr(N_{\ell+1}^t = Y_b \mid P_\ell^t = \rho, \mathcal{E}^t, Q^t = q) = 1.0$.

## 3.2 BELIEF STATE INITIALIZATION

The initial belief state begins with $B_Q^1(q) = \Pr(Q^0 = q)$, easily obtained from the prior probability function $\pi_0(q)$. We can then work top down, computing the probability for $B_N^1(1, S, q)$, $B_P^1(1, \langle a, b\rangle, q)$, $B_N^1(2, X, 1), \ldots, B_\Sigma^1(x, q)$. At each step, we compute production and symbol probabilities using the generative method used in computing the probability of the sample parse tree from Section 2.2.

### 3.3 EXPLANATION PROBABILITIES

Given a new observation at time $t$, of the form $Q^{t-1} \in R^{t-1}$, we can easily compute the probability of the individual state instantiations in $R^{t-1}$ using $\pi_1$, $B_Q^{t-1}$, and $B_\Sigma^{t-1}$. With the example observations of Figure 2, we would first compute the probability of the observed state $Q^1$ given the initial state $Q^0$ and the possible terminal symbols. Once we had these probabilities, we can marginalize over the set of terminal symbols to determine the probability of the observed $Q^1$ given only $Q^0$. In general, the time complexity of computing this probability distribution is $O(|R|^2|\Sigma|)$.

We can then proceed bottom-up through the subplan hierarchy to compute the probability of the evidence conditioned on the possible states of the nonterminal symbol nodes, similar to a generalization of the transition probability function $\pi_1$. These probability values are reused many times in subsequent computations within the current time slice. We can compute such probabilities recursively by starting with the base definition of $\pi_1$ over all terminal symbols $x \in \Sigma$. We then proceed up through the hierarchy, where at each level, we compute the probability of all state transitions (consistent with our prior beliefs and new observations) given the possible nonterminal symbols. For each symbol, we can compute this transition probability by examining all of the possible expansions (based on our prior beliefs) in the context of the transition probabilities of the symbols on the right-hand side (computed in previous dynamic programming passes). If $m$ is the maximum production length, and $d$ is the depth of the hierarchy, this dynamic programming phase takes time $O(|R|^2|P|md)$.

We can use the dynamic programming results to obtain posterior probability distribution over symbols and productions at time $t-1$ conditioned on evidence up to and including time $t$, useful for answering explanation queries. The computation requires only the constant-time combination of our prior beliefs over symbols with the transition probabilities over these symbols.

### 3.4 PREDICTION PROBABILITIES

After completing the explanation phase, we compute prediction probabilities for time $t$ using the posterior probabilities over the variables at time $t-1$. We marginalize over the two possible termination cases for $t-1$, i.e., either an expansion terminated or it has not. If it has, then we choose a whole new production at time $t$ using the production probability functions. If the expansion has not terminated at time $t-1$, then we continue the expansion at time $t$. If the child symbol's expansion has terminated at time $t-1$, then we deterministically move to the next symbol on the right-hand side at time $t$. Otherwise, we continue expanding the same child symbol. We must then marginalize over the possible states, so the time complexity of computing all of the prediction probabilities is $O(|R|d|P|m)$.

### 3.5 COMPUTATION OF NEW BELIEF STATE

The prediction phase specifies the symbol and production components of the new belief state $B^t$. It is straightforward, from the definition of termination, to compute the required probability of termination given a particular symbol in a single bottom-up pass through the symbol and production probabilities at each level of the hierarchy. We can then marginalize this result to compute the termination probability independent of symbol. We can compute these probabilities in a single bottom-up pass through the hierarchy requiring time $O(|R|d|P|)$.

### 3.6 EVALUATION OF PSDG INFERENCE

Overall, these algorithms compute prediction and explanation probabilities over the low-level actions, complex plans, and intermediate plan states. In addition, the belief state continually updates its distribution over the unobserved state variables, allowing a recognizing agent to reason about another agent's mental state. The PSDG inference algorithms thus support many of the queries desired by recognizing agents in multiagent environments.

However, the algorithms cannot exploit direct evidence about plans. Evidence about subplan choices usually comes in the form of explicit communication. For instance, a driver in a convoy may radio its intended lane change to the other drivers. Such evidence would introduce new dependencies to the belief state structure that are likely to greatly increase the complexity of the inference algorithms.

The overall inference algorithms for a single time step have time complexity $O(|R| \cdot |\Sigma| \cdot |Q| + |R|^2 d|P|m)$. If we do not compute a probability distribution over future state $Q^t$ (the distribution is not necessary for answering queries about only plans), the time complexity is $O(|R|^2 d|P|m)$. An entire inference cycle (explanation, prediction, and belief update) takes 1 CPU second for the full traffic monitoring PSDG (where $|Q| \approx 6 \times 10^{14}$, $|R| = 18$, $|\Sigma| = 7$, $d = 6$, $|P| = 37$, and $m = 3$), with the inference algorithms running on a SUN Sparc machine. The inference for the air combat PSDG (where $|Q| = 1150$, $|R| = 1$, $|\Sigma| = 3$, $d = 6$, $|P| = 34$, and $m = 3$) was even faster due to the complete observability of the state variables in that domain.

Although the specialized algorithms save considerably over the DBN algorithms, the time and space complexity is still quadratic in the number of instantiations of the unobserved state variables. This cost is potentially prohibitive, since the number of such state instantiations grows exponentially with the number of unobserved state variables. This complexity is clearly the limiting factor when determining the tractability of the PSDG approach in a given domain.

## 4 CONCLUSION

The assumptions of the PSDG model and inference algorithms sacrifice the generality of some existing probabilistic languages (Koller et al., 1997; Goldman et al., 1999). However, the restrictions of the PSDG model produce the independence properties that the algorithms exploit for practical inference. If we relax these restrictions (e.g., state transition probabilities depending on nonterminal symbols as well), we can no longer partition the belief state along the different levels of the hierarchy. Even within the existing restrictions, in domains with more complex models of unobservable mental states, the complexity of inference could be prohibitive. One potential solution is the use of approximation methods used with similar dynamic belief models (Boyen & Koller, 1998; Ghahramani & Jordan, 1997).

Learning algorithms, analogous to those for PCFGs, could potentially automatically generate PSDG productions, states, and probabilities based on labeled parse trees. Such learning algorithms would reduce the effort required in domain specification, as well as potentially supporting dynamic PSDGs that could adapt to changes in an agent's behavior. However, in domains where the observed agent's behavior depends significantly on the recognizing agent's decisions (as in the clearly adversarial domain of air combat), even such a dynamic PSDG specification would be too weak to perform the reflexive modeling required.

The PSDG language contributes a new representation that exploits the partition between plans and state that exists in most plan-recognition domains. We successfully created PSDG models of planning agents in two domains, one requiring creation of a new specification from scratch and the other requiring translation of an existing specification into the PSDG language. We were able to develop algorithms for automatic generation of a DBN representation of a PSDG domain model, but the resulting networks were impractical for inference. We then designed specialized algorithms that used a compact belief state to summarize the entire sequence of observations while incurring time and space complexity costs that are sublinear in the space of possible plan instantiations. Therefore, the PSDG language model supports practical probabilistic plan recognition in domains where existing languages do not.